\documentclass[10pt,onecolumn,letterpaper]{article}

\usepackage{iccv}
\usepackage{times}
\usepackage{epsfig}
\usepackage{graphicx}
\usepackage{amsmath}
\usepackage{amssymb}
\usepackage{graphicx}
\usepackage{amsmath}
\usepackage{amssymb}
\usepackage{booktabs}
\usepackage{cite}
\usepackage{booktabs}
\usepackage{multirow}
\usepackage[table,xcdraw]{xcolor}

\usepackage[pagebackref=true,breaklinks=true,letterpaper=true,colorlinks,bookmarks=false]{hyperref}

\iccvfinalcopy 

\begin{document}


\title{Can SAM Segment Anything? \\ {\small -When SAM Meets Camouflaged Object Detection}}

\author{Lv Tang$^1$ \quad Haoke Xiao$^3$  \quad Bo Li$^2$\thanks{Corresponding author} \\
{$^1$Independent Researcher, Beijing, China} \\
{$^2$Youtu Lab, Tencent, Shanghai, China} \\
{$^3$Xiamen University, Xiamen, China} \\
{\tt\small luckybird1994@gmail.com}, {\tt\small hk.xiao.me@gmail.com}, 
{\tt\small libraboli@tencent.com}}
\maketitle

\begin{abstract}
SAM is a segmentation model recently released by Meta AI Research and has been gaining attention quickly due to its impressive performance in generic object segmentation. However, its ability to generalize to specific scenes such as camouflaged scenes is still unknown. Camouflaged object detection (COD) involves identifying objects that are seamlessly integrated into their surroundings and has numerous practical applications in fields such as medicine, art, and agriculture. In this study, we try to ask if SAM can address the COD task and evaluate the performance of SAM on the COD benchmark by employing maximum segmentation evaluation and camouflage location evaluation. We also compare SAM's performance with 22 state-of-the-art COD methods. Our results indicate that while SAM shows promise in generic object segmentation, its performance on the COD task is limited. This presents an opportunity for further research to explore how to build a stronger SAM that may address the COD task. The results of this paper are provided in \url{https://github.com/luckybird1994/SAMCOD}.
\end{abstract}

\section{Introduction}
Large language models (LLMs), also known as foundation models \cite{DBLP:conf/nips/BrownMRSKDNSSAA20,DBLP:journals/corr/abs-2108-07258}, have achieved significant performance on various natural language processing (NLP) tasks. Recently, OpenAI released a chatbot service called ChatGPT, which is a variant of the Generative Pre-trained Transformers (GPT) family. Thanks to its user-friendly interface and exceptional performance, ChatGPT has attracted over 100 million users within just two months.

Unlike NLP, there is still much to explore in the area of foundation models in computer vision (CV). While some models like CLIP \cite{DBLP:conf/icml/RadfordKHRGASAM21} and ALIGN \cite{DBLP:conf/icml/JiaYXCPPLSLD21} have been designed, many problems in CV remain unsolved. Recently, Meta AI Research developed a foundation model called SAM \cite{kirillov2023segment} for image segmentation. SAM is a promptable model that is pre-trained on a broad dataset using a task that enables powerful generalization. The remarkable performance of SAM in image segmentation has astonished researchers in CV. Some have even gone as far as to exclaim that ``CV does not exist", a reference to a famous line from the science fiction novel Three-Body.

In this technical report, we evaluate the performance of SAM when applied to the task of camouflaged object detection (COD). The goal of COD is to accurately identify objects that are seamlessly blended into their background. This task has significant scientific and practical value in various fields, such as medicine \cite{fan2020pranet,DBLP:journals/tmi/FanZJZCFSS20}, agriculture \cite{DBLP:journals/cea/ChengZCWY17,DBLP:conf/cvpr/WuZLCY19}, and art \cite{DBLP:journals/tog/ChuHMCWL10}. To evaluate SAM's performance on COD benchmarks, we follow two rules: 1) Can SAM accurately segment camouflaged objects? and 2) Can SAM accurately locate camouflaged objects? Therefore, we use maximum segmentation evaluation and camouflage location evaluation to assess the performance of SAM. Our comparison results between SAM and 22 state-of-the-art COD methods reveal that SAM has limitations when it comes to COD tasks. We hope that our findings can inspire and give confidence to researchers in the COD field, despite SAM's limitations in addressing the COD task. Furthermore, we believe that this presents an opportunity for further research to explore how to effectively apply SAM to the COD task.

\section{Experiments}\label{sec4}
\subsection{Datasets, Comparison and Metrics}
\textbf{Datasets.} To validate the effectiveness of SAM, we evaluate its performance on three benchmark datasets, containing CAMO~\cite{DBLP:journals/cviu/LeNNTS19}, COD10K~\cite{DBLP:conf/cvpr/FanJSCS020} and NC4K~\cite{DBLP:conf/cvpr/Lv0DLLBF21}.

\textbf{Comparison Methods.} We compare SAM with 22 state-of-the-art COD methods, including the SINet~\cite{DBLP:conf/cvpr/FanJSCS020}, C2FNet~\cite{DBLP:conf/ijcai/SunCZZL21}, LSR~\cite{DBLP:conf/cvpr/Lv0DLLBF21}, PFNet~\cite{DBLP:conf/cvpr/MeiJW0WF21}, MGL~\cite{DBLP:conf/cvpr/ZhaiL0C0F21}, JCOD~\cite{DBLP:conf/cvpr/Li0LL0D21}, TANet~\cite{DBLP:conf/aaai/ZhuZZL21}, BGNet~\cite{DBLP:conf/ijcai/SunWCX22}, FDCOD~\cite{DBLP:conf/cvpr/ZhongLTKWD22}, SegMaR~\cite{DBLP:conf/cvpr/0001YL0LL22}, ZoomNet~\cite{DBLP:conf/cvpr/PangZXZL22}, BSANet~\cite{DBLP:conf/aaai/ZhuL0YLCWQ22}, SINetV2~\cite{DBLP:journals/pami/FanJCS22}, FAPNet~\cite{DBLP:journals/tip/ZhouZGYZ22}, the extension version of C2FNet~\cite{DBLP:journals/tcsv/ChenLSJWZ22}, DGNet~\cite{DBLP:journals/ijautcomp/JiFCDLG23}, CubeNet~\cite{DBLP:journals/pr/ZhugeLGCC22}, ERRNet~\cite{DBLP:journals/pr/JiZZF22}, VST~\cite{DBLP:conf/iccv/LiuZW0H21}, UGTR~\cite{DBLP:conf/iccv/0054Z00L0F21}, ICON~\cite{DBLP:journals/pami/ZhugeFLZXS23} and TPRNet~\cite{2022TPRNet}.

\textbf{Evaluation Metrics.} We use six widely used metrics: structure-measure ($S_\alpha$)~\cite{DBLP:conf/iccv/FanCLLB17}, mean E-measure ($E_\phi$)~\cite{21Fan_HybridLoss}, F-measure ($F_\beta$), weighted F-measure ($F^w_\beta$), max F-measure ($F^{max}_\beta$)~\cite{DBLP:conf/cvpr/MargolinZT14}, and mean absolute error (MAE).

\begin{table}[!htbp]
\centering
\caption{Comparison Results between SAM and 22 state-of-the-art COD methods.}
\scalebox{0.55}{
\begin{tabular}{@{}cccccccccccccccccccc@{}}
\toprule
\multicolumn{1}{c|}{}                                             & \multicolumn{1}{c|}{}                                                  & \multicolumn{6}{c|}{\textbf{CAMO-Test (250 images)}}                                                                                                                                                                                 & \multicolumn{6}{c|}{COD10K-Test (2026 images)}                                                                                                                                                               & \multicolumn{6}{c}{\textbf{NC4K (4121 images)}}                                                                                                                                         \\ \cmidrule(l){3-20} 
\multicolumn{1}{c|}{\multirow{-2}{*}{Methods}}                    & \multicolumn{1}{c|}{\multirow{-2}{*}{\textbf{Pub.}}}                   & $S_\alpha$                                           & $E_\phi$                     & $F^w_\beta$                  & $F_\beta$                    & $F^{max}_\beta$              & \multicolumn{1}{c|}{MAE}                          & $S_\alpha$                   & $E_\phi$                     & $F^w_\beta$                  & $F_\beta$                    & $F^{max}_\beta$              & \multicolumn{1}{c|}{MAE}                          & $S_\alpha$                   & $E_\phi$                     & $F^w_\beta$                  & $F_\beta$                    & $F^{max}_\beta$              & MAE                          \\ \midrule
\multicolumn{20}{c}{CNN-Based Models}                                                                                                                                                                                                                                                                                                                                                                                                                                                                                                                                                                                                                                                                                                                                                      \\ \midrule
\multicolumn{1}{c|}{SINet}                                        & \multicolumn{1}{c|}{CVPR'20}                                           & 0.751                                                & 0.771                        & 0.606                        & 0.675                        & 0.706                        & \multicolumn{1}{c|}{0.100}                        & 0.771                        & 0.806                        & 0.551                        & 0.634                        & 0.676                        & \multicolumn{1}{c|}{0.051}                        & 0.808                        & 0.871                        & 0.723                        & 0.769                        & 0.775                        & 0.058                        \\
\multicolumn{1}{c|}{C2FNet}                                       & \multicolumn{1}{c|}{IJCAI'21}                                          & 0.796                                                & 0.854                        & 0.719                        & 0.762                        & 0.771                        & \multicolumn{1}{c|}{0.080}                        & 0.812                        & 0.890                        & 0.686                        & 0.723                        & 0.743                        & \multicolumn{1}{c|}{0.036}                        & 0.838                        & 0.897                        & 0.762                        & 0.795                        & 0.810                        & 0.049                        \\
\multicolumn{1}{c|}{LSR}                                          & \multicolumn{1}{c|}{CVPR'21}                                           & 0.787                                                & 0.838                        & 0.696                        & 0.744                        & 0.753                        & \multicolumn{1}{c|}{0.080}                        & 0.804                        & 0.880                        & 0.673                        & 0.715                        & 0.732                        & \multicolumn{1}{c|}{0.037}                        & 0.840                        & 0.895                        & 0.766                        & 0.804                        & 0.815                        & 0.048                        \\
\multicolumn{1}{c|}{PFNet}                                        & \multicolumn{1}{c|}{CVPR'21}                                           & 0.782                                                & 0.841                        & 0.695                        & 0.746                        & 0.758                        & \multicolumn{1}{c|}{0.085}                        & 0.800                        & 0.877                        & 0.660                        & 0.701                        & 0.725                        & \multicolumn{1}{c|}{0.040}                        & 0.829                        & 0.887                        & 0.745                        & 0.784                        & 0.799                        & 0.053                        \\
\multicolumn{1}{c|}{MGL}                                          & \multicolumn{1}{c|}{CVPR'21}                                           & 0.775                                                & 0.812                        & 0.673                        & 0.726                        & 0.740                        & \multicolumn{1}{c|}{0.088}                        & 0.814                        & 0.851                        & 0.666                        & 0.710                        & 0.738                        & \multicolumn{1}{c|}{0.035}                        & 0.833                        & 0.867                        & 0.739                        & 0.782                        & 0.800                        & 0.053                        \\
\multicolumn{1}{c|}{JCOD}                                         & \multicolumn{1}{c|}{CVPR'21}                                           & 0.800                                                & 0.859                        & 0.728                        & 0.772                        & 0.779                        & \multicolumn{1}{c|}{0.073}                        & 0.809                        & 0.884                        & 0.684                        & 0.721                        & 0.738                        & \multicolumn{1}{c|}{0.035}                        & 0.842                        & 0.898                        & 0.771                        & 0.806                        & 0.816                        & 0.047                        \\
\multicolumn{1}{c|}{TANet}                                        & \multicolumn{1}{c|}{AAAI'21}                                           & 0.781                                                & 0.847                        & 0.678                        & -                            & -                            & \multicolumn{1}{c|}{0.087}                        & 0.793                        & 0.848                        & 0.635                        & -                            & -                            & \multicolumn{1}{c|}{0.043}                        & -                            & -                            & -                            & -                            & -                            & -                            \\
\multicolumn{1}{c|}{BGNet}                                        & \multicolumn{1}{c|}{IJCAI'22}                                          & 0.813                                                & 0.870                        & 0.749                        & 0.789                        & 0.799                        & \multicolumn{1}{c|}{0.073}                        & 0.831                        & {\color[HTML]{330001} 0.901} & 0.722                        & 0.753                        & 0.774                        & \multicolumn{1}{c|}{0.033}                        & 0.851                        & 0.907                        & {\color[HTML]{FE0000} 0.788} & 0.820                        & {\color[HTML]{FE0000} 0.833} & 0.044                        \\
\multicolumn{1}{c|}{FDCOD}                                        & \multicolumn{1}{c|}{CVPR'22}                                           & 0.828                                                & 0.883                        & 0.748                        & 0.781                        & 0.804                        & \multicolumn{1}{c|}{0.068}                        & 0.832                        & {\color[HTML]{FE0000} 0.907} & 0.706                        & 0.733                        & 0.776                        & \multicolumn{1}{c|}{0.033}                        & 0.834                        & 0.893                        & 0.750                        & {\color[HTML]{FE0000} 0.784} & 0.804                        & 0.051                        \\
\multicolumn{1}{c|}{SegMaR}                                       & \multicolumn{1}{c|}{CVPR'22}                                           & 0.815                                                & 0.874                        & 0.753                        & 0.795                        & 0.803                        & \multicolumn{1}{c|}{0.071}                        & 0.833                        & 0.899                        & 0.724                        & 0.757                        & 0.774                        & \multicolumn{1}{c|}{0.034}                        & 0.841                        & 0.896                        & 0.781                        & 0.821                        & 0.826                        & 0.046                        \\
\multicolumn{1}{c|}{ZoomNet}                                      & \multicolumn{1}{c|}{CVPR'22}                                           & 0.820                                                & 0.877                        & 0.752                        & 0.794                        & 0.805                        & \multicolumn{1}{c|}{0.066}                        & {\color[HTML]{FE0000} 0.838} & 0.888                        & {\color[HTML]{FE0000} 0.729} & {\color[HTML]{FE0000} 0.766} & {\color[HTML]{FE0000} 0.780} & \multicolumn{1}{c|}{{\color[HTML]{FE0000} 0.029}} & 0.853                        & 0.896                        & 0.784                        & 0.818                        & 0.828                        & 0.043                        \\
\multicolumn{1}{c|}{BSANet}                                       & \multicolumn{1}{c|}{AAAI'22}                                           & 0.794                                                & 0.851                        & 0.717                        & 0.763                        & 0.770                        & \multicolumn{1}{c|}{0.079}                        & 0.817                        & 0.891                        & 0.699                        & 0.738                        & 0.753                        & \multicolumn{1}{c|}{0.034}                        & 0.841                        & 0.897                        & 0.771                        & 0.808                        & 0.817                        & 0.048                        \\ \midrule
\multicolumn{1}{c|}{SINetV2}                                      & \multicolumn{1}{c|}{PAMI'22}                                           & 0.822                                                & 0.882                        & 0.743                        & 0.782                        & 0.801                        & \multicolumn{1}{c|}{0.070}                        & 0.815                        & 0.887                        & 0.680                        & 0.718                        & 0.752                        & \multicolumn{1}{c|}{0.037}                        & 0.847                        & 0.903                        & 0.770                        & 0.805                        & 0.823                        & 0.048                        \\
\multicolumn{1}{c|}{FAPNet}                                       & \multicolumn{1}{c|}{TIP'22}                                            & 0.815                                                & 0.865                        & 0.734                        & 0.776                        & 0.792                        & \multicolumn{1}{c|}{0.076}                        & 0.822                        & 0.887                        & 0.694                        & 0.731                        & 0.758                        & \multicolumn{1}{c|}{0.036}                        & 0.851                        & 0.899                        & 0.775                        & 0.810                        & 0.825                        & 0.046                        \\
\multicolumn{1}{c|}{C2FNet-ext}                                   & \multicolumn{1}{c|}{TCSVT'22}                                          & 0.799                                                & 0.859                        & 0.730                        & 0.770                        & 0.779                        & \multicolumn{1}{c|}{0.077}                        & 0.811                        & 0.887                        & 0.691                        & 0.725                        & 0.742                        & \multicolumn{1}{c|}{0.036}                        & 0.840                        & 0.896                        & 0.770                        & 0.802                        & 0.814                        & 0.048                        \\
\multicolumn{1}{c|}{DGNet}                                        & \multicolumn{1}{c|}{MIR'22}                                            & 0.839                                                & {\color[HTML]{FE0000} 0.900} & 0.768                        & {\color[HTML]{FE0000} 0.806} & 0.822                        & \multicolumn{1}{c|}{{\color[HTML]{FE0000} 0.057}} & 0.822                        & 0.896                        & 0.693                        & 0.728                        & 0.759                        & \multicolumn{1}{c|}{0.033}                        & {\color[HTML]{FE0000} 0.854} & 0.909                        & 0.783                        & 0.813                        & 0.830                        & {\color[HTML]{FE0000} 0.043} \\
\multicolumn{1}{c|}{CubeNet}                                      & \multicolumn{1}{c|}{PR'22}                                             & 0.788                                                & 0.838                        & 0.682                        & 0.732                        & 0.750                        & \multicolumn{1}{c|}{0.085}                        & 0.795                        & 0.865                        & 0.643                        & 0.692                        & 0.715                        & \multicolumn{1}{c|}{0.041}                        & -                            & -                            & -                            & -                            & -                            & -                            \\
\multicolumn{1}{c|}{ERRNet}                                       & \multicolumn{1}{c|}{PR'22}                                             & 0.779                                                & 0.842                        & 0.679                        & 0.729                        & 0.742                        & \multicolumn{1}{c|}{0.085}                        & 0.786                        & 0.867                        & 0.630                        & 0.675                        & 0.702                        & \multicolumn{1}{c|}{0.043}                        & 0.827                        & 0.887                        & 0.737                        & 0.778                        & 0.794                        & 0.054                        \\ \midrule
\multicolumn{20}{c}{Transformer-Based Models}                                                                                                                                                                                                                                                                                                                                                                                                                                                                                                                                                                                                                                                                                                                                              \\ \midrule
\multicolumn{1}{c|}{VST}                                          & \multicolumn{1}{c|}{ICCV'21}                                           & 0.807                                                & 0.848                        & 0.713                        & 0.758                        & 0.777                        & \multicolumn{1}{c|}{0.081}                        & 0.820                        & 0.879                        & 0.698                        & 0.738                        & 0.754                        & \multicolumn{1}{c|}{0.037}                        & 0.845                        & 0.893                        & 0.767                        & 0.804                        & 0.817                        & 0.048                        \\
\multicolumn{1}{c|}{UGTR}                                         & \multicolumn{1}{c|}{ICCV'21}                                           & 0.785                                                & 0.822                        & 0.685                        & 0.737                        & 0.753                        & \multicolumn{1}{c|}{0.086}                        & 0.818                        & 0.852                        & 0.667                        & 0.712                        & 0.742                        & \multicolumn{1}{c|}{0.035}                        & 0.839                        & 0.874                        & 0.746                        & 0.787                        & 0.807                        & 0.052                        \\
\multicolumn{1}{c|}{ICON}                                         & \multicolumn{1}{c|}{PAMI‘22}                                           & {\color[HTML]{FE0000} 0.840}                         & 0.894                        & {\color[HTML]{FE0000} 0.769} & 0.796                        & {\color[HTML]{FE0000} 0.824} & \multicolumn{1}{c|}{0.058}                        & 0.818                        & 0.904                        & 0.688                        & 0.717                        & 0.756                        & \multicolumn{1}{c|}{0.033}                        & 0.847                        & {\color[HTML]{FE0000} 0.911} & 0.784                        & 0.697                        & 0.817                        & 0.045                        \\
\multicolumn{1}{c|}{TPRNet}                                       & \multicolumn{1}{c|}{TVCJ'22}                                           & 0.807                                                & 0.861                        & 0.725                        & 0.772                        & 0.785                        & \multicolumn{1}{c|}{0.074}                        & 0.817                        & 0.887                        & 0.683                        & 0.724                        & 0.748                        & \multicolumn{1}{c|}{0.036}                        & 0.846                        & 0.898                        & 0.768                        & 0.805                        & 0.820                        & 0.048                        \\
\multicolumn{1}{c|}{{\color[HTML]{333333} }}                      & \multicolumn{1}{c|}{{\color[HTML]{333333} }}                           & \cellcolor[HTML]{FFFFFF}{\color[HTML]{6200C9} 0.684} & {\color[HTML]{6200C9} 0.687} & {\color[HTML]{6200C9} 0.606} & {\color[HTML]{6200C9} 0.680} & {\color[HTML]{6200C9} 0.681} & \multicolumn{1}{c|}{{\color[HTML]{6200C9} 0.132}} & {\color[HTML]{6200C9} 0.783} & {\color[HTML]{6200C9} 0.798} & {\color[HTML]{6200C9} 0.701} & {\color[HTML]{6200C9} 0.756} & {\color[HTML]{6200C9} 0.758} & \multicolumn{1}{c|}{{\color[HTML]{6200C9} 0.050}} & {\color[HTML]{6200C9} 0.767} & {\color[HTML]{6200C9} 0.776} & {\color[HTML]{6200C9} 0.696} & {\color[HTML]{6200C9} 0.752} & {\color[HTML]{6200C9} 0.754} & {\color[HTML]{6200C9} 0.078} \\
\multicolumn{1}{c|}{\multirow{-2}{*}{{SAM}}} & \multicolumn{1}{c|}{\multirow{-2}{*}{{arXiv'23}}} & \cellcolor[HTML]{FFFFFF}{\color[HTML]{6200C9} 19\%}  & {\color[HTML]{6200C9} 24\%}  & {\color[HTML]{6200C9} 21\%}  & {\color[HTML]{6200C9} 16\%}  & {\color[HTML]{6200C9} 17\%}  & \multicolumn{1}{c|}{{\color[HTML]{6200C9} 56\%}}  & {\color[HTML]{6200C9} 7\%}   & {\color[HTML]{6200C9} 12\%}  & {\color[HTML]{6200C9} 4\%}   & {\color[HTML]{6200C9} 1\%}   & {\color[HTML]{6200C9} 3\%}   & \multicolumn{1}{c|}{{\color[HTML]{6200C9} 42\%}}  & {\color[HTML]{6200C9} 10\%}  & {\color[HTML]{6200C9} 15\%}  & {\color[HTML]{6200C9} 12\%}  & {\color[HTML]{6200C9} 4\%}   & {\color[HTML]{6200C9} 9\%}   & {\color[HTML]{6200C9} 45\%}  \\ \bottomrule
\end{tabular}}
\label{table1}
\end{table}

\begin{figure*}[!htbp]
    \centering
    \includegraphics[width=0.75\linewidth]{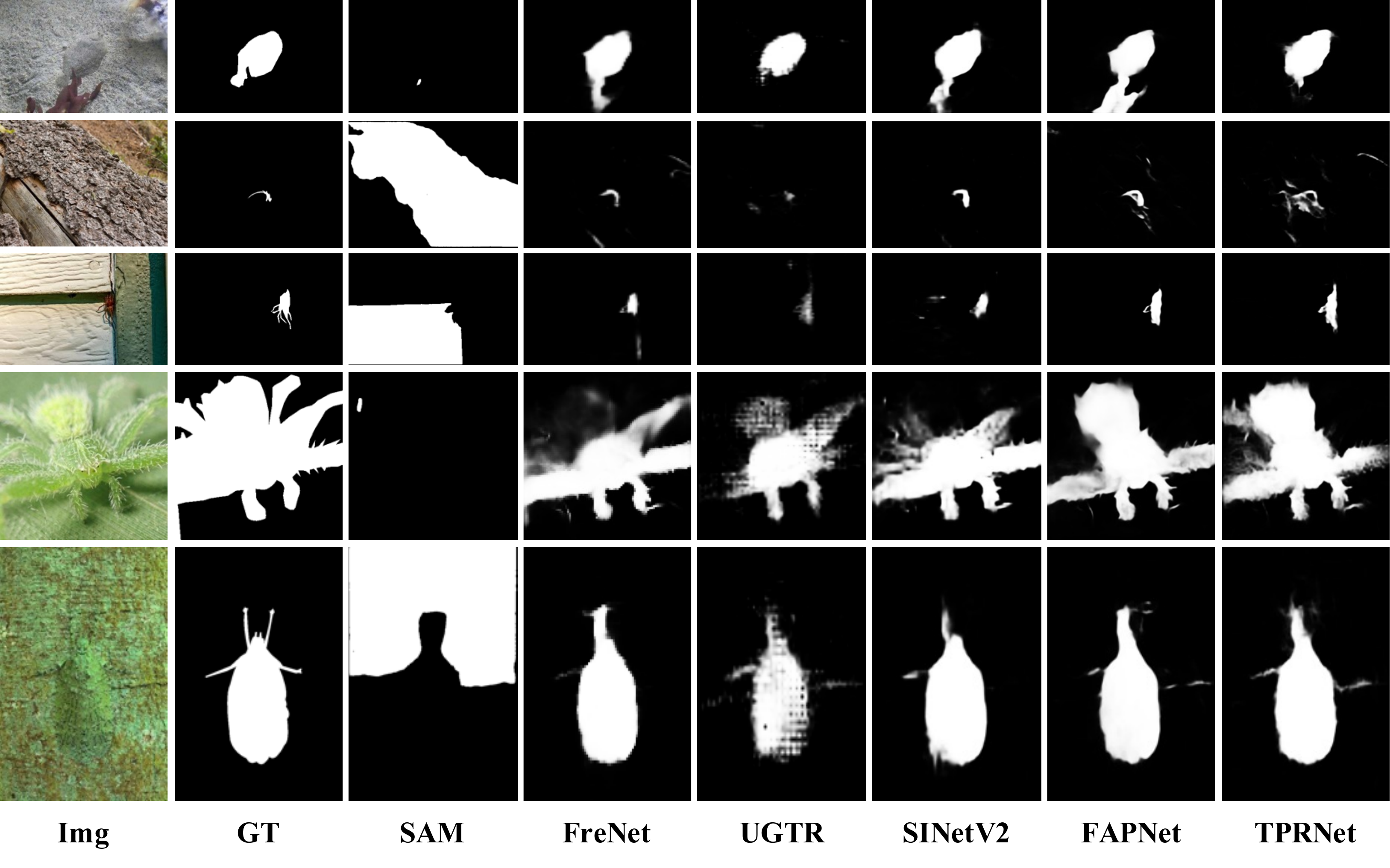}
    \caption{Some bad cases of SAM.}
    \label{bad}
\end{figure*}

\subsection{Maximum Segmentation Evaluation}

In general, when evaluating the performance of a COD method on a certain benchmark, the performance of each predicted map is calculated first, and then they are averaged. This process for calculating the $F_\beta$ score can be written as follows:
\begin{equation}
    F_\beta = \frac{1}{N}\sum_{i=1}^{i=N}F_\beta^i,
\end{equation}
where $N$ is the total number of a dataset. 

SAM generates multiple binary maps for each image, but not all of them contain the camouflaged object. To evaluate SAM's upper segmentation performance, we propose the maximum segmentation evaluation. For a given image, we calculate the mean F-measure for all predicted maps and select the one with the highest F-measure value as SAM's predicted map. Consequently, we express SAM's $F_\beta$ on a particular dataset as:
\begin{equation}
    \begin{aligned}
        HF^i &= max\{F^i_1,F^i_2,...,F^i_m\}, \\
        F_\beta &= \frac{1}{N}\sum_{i=1}^{i=N}HF^i,
    \end{aligned}
\end{equation}
where $m$ is the total number of predicted binary maps for one image, and $F^i_m$ is the mean F-measure of one binary map. Then, we can calculate the remaining metrics for the SAM. As shown in Table. \ref{table1} and Fig. \ref{bad}, the performance of SAM is still worse than state-of-the-art COD methods.

\subsection{General Location Evaluation}

\begin{table}[!htbp]
\centering
\caption{The ratios of predicted binary maps that exceed a certain $F_\beta$ threshold}
\begin{tabular}{@{}c|ccccc@{}}
\toprule
Thresholds & 0.2    & 0.3    & 0.4    & 0.5    & 0.7    \\ \midrule
CAMO       & 2.55\% & 1.83\% & 1.39\% & 1.15\% & 0.82\% \\
COD10K     & 2.32\% & 1.80\% & 1.48\% & 1.26\% & 0.93\% \\
NC4K       & 2.94\% & 2.22\% & 1.82\% & 1.53\% & 1.13\% \\ \bottomrule
\end{tabular}
\label{table2}
\end{table}

\begin{figure*}[!htbp]
    \centering
    \includegraphics[width=0.9\linewidth]{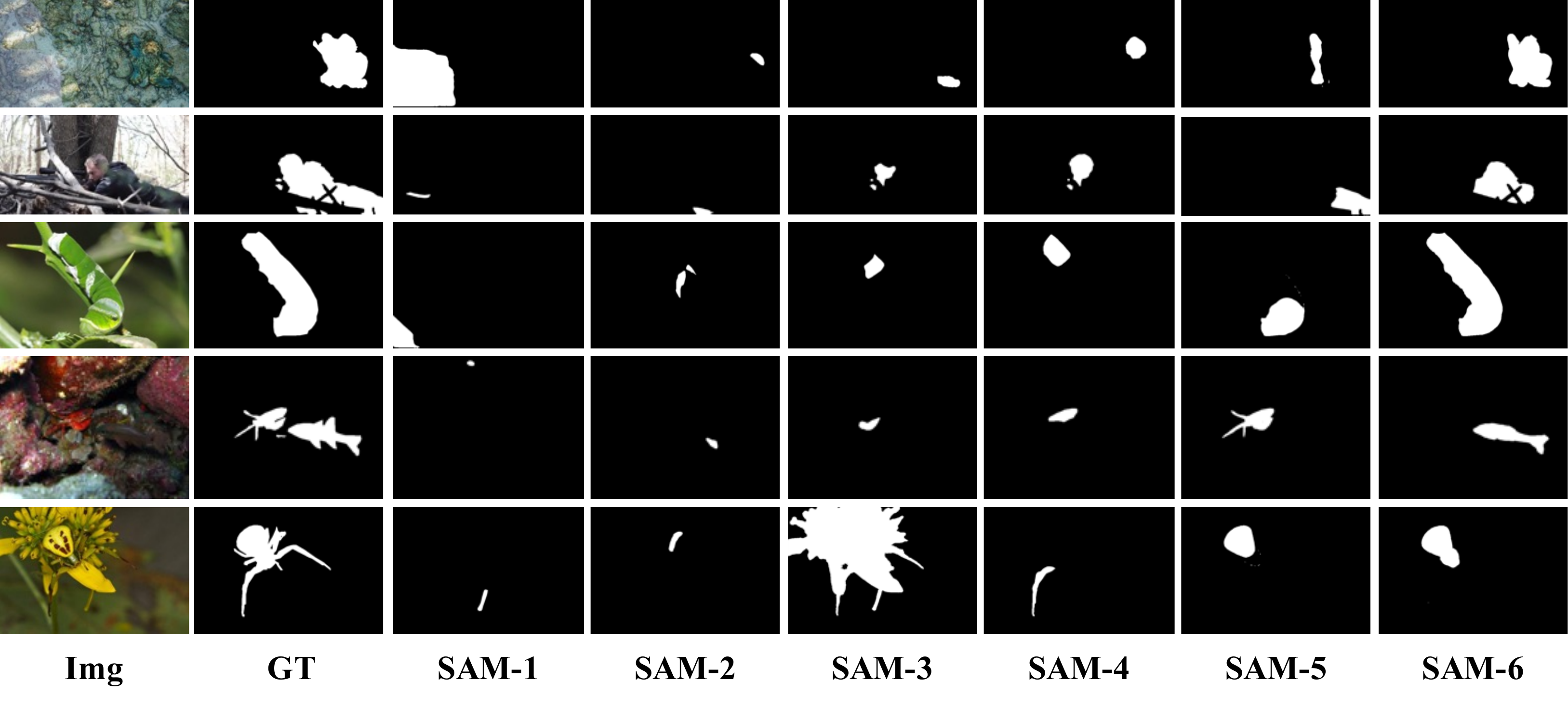}
    \caption{The location ability of the SAM.}
    \label{location}
\end{figure*}

SAM generates multiple binary maps for each image, and we have designed the general location evaluation to assess how accurately SAM can detect the camouflaged object. Specifically, for a given image, we calculate the $F_\beta$ for all predicted maps and then calculate the ratios of maps that exceed a certain $F_\beta$ threshold. As shown in Table. \ref{table2} and Fig. \ref{location}, the location performance of SAM is needed to be further improved. 

\begin{figure*}
    \centering
    \includegraphics[width=0.75\linewidth]{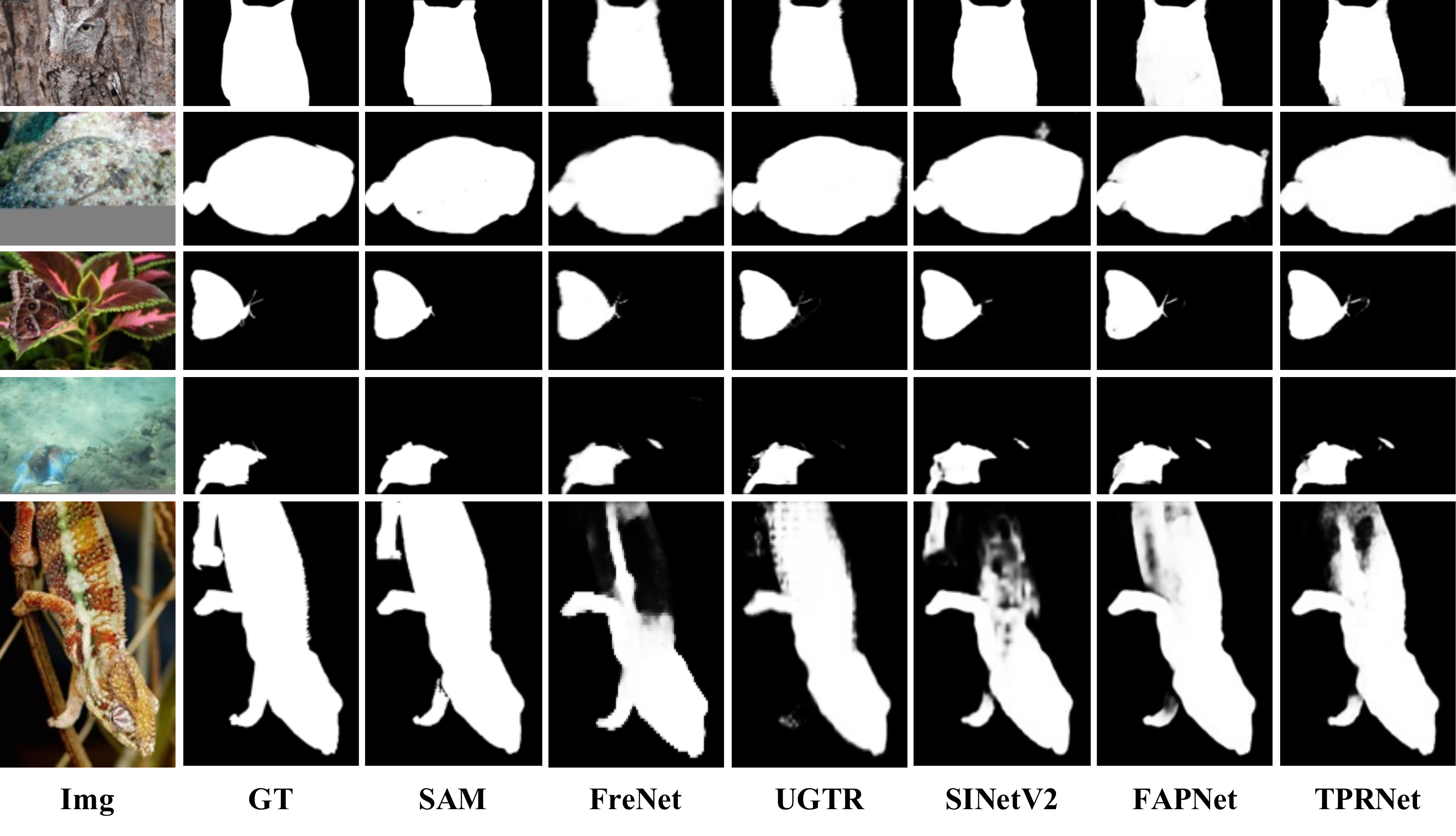}
    \caption{Some good cases of SAM.}
    \label{good}
\end{figure*}

\section{Conclusion}
This paper presents a preliminary evaluation of SAM's performance when applied to the COD task. While we acknowledge the advancement of large CV foundation models, our experiments show that there is still room for improvement in SAM's performance on this specific task. Although SAM's performance lags behind some state-of-the-art COD methods, Table. \ref{table1} and Fig. \ref{good} show that SAM can achieve a noteworthy performance compared to some methods published in 2021. We hope that our evaluation and analysis can provide insights for future research, such as designing a stronger foundation model or improving the network architecture of SAM.

\clearpage

{\small
\bibliographystyle{ieee_fullname}
\bibliography{egbib}
}

\end{document}